\documentclass[10pt,twocolumn,letterpaper]{article}

\usepackage{3dv}
\makeatletter
\@namedef{ver@everyshi.sty}{}
\makeatother
\usepackage{times}
\usepackage{epsfig}
\usepackage{amsmath}
\usepackage{amssymb}
\usepackage{booktabs}

\usepackage{units}
\usepackage[nolist]{acronym}
\usepackage{pgfplots}
\usepackage{tikz} 
\pgfplotsset{compat=newest}
\usepackage{textcomp}
\usepackage{blindtext}
\usepackage{subcaption}
\usepackage{multirow}
\usepackage{array}
\usepackage{calc}
\usepackage{booktabs}
\usepackage{graphicx}

\usepackage[pagebackref=true,breaklinks=true,letterpaper=true,colorlinks,bookmarks=false]{hyperref}

\threedvfinalcopy % *** Uncomment this line for the final submission

\def\threedvPaperID{17} % *** Enter the 3DV Paper ID here

\ifthreedvfinal\fi

\newcommand{\arcsec}[1]{$\text{#1}"$}

    \def\addlegendimage{\csname pgfplots@addlegendimage\endcsname}
\begin{acronym}
 \acro{ALE}{average log error}
 \acro{ARD}{absolute relative distance}
 \acro{CNN}{convolutional neural network}
 \acro{GPS}{global positioning system}
 \acro{ICP}{iterative closest point}
 \acro{IMU}{inertial measurement unit}
 \acro{logRMSE}{logarithmic root-mean-squared error}
 \acro{MAE}{mean-absolute error}
 \acro{ppm}{parts per million}
 \acro{PPS}{pulse per second}
 \acro{PSNR}{peak signal-to-noise ratio}
 \acro{rPSNR}{relative peak signal-to-noise ratio}
 \acro{RMSE}{root-mean-squared error}
 \acro{ROS}{robot operating system}
 \acro{SfM}{structure from motion}
 \acro{SGM}{semi-global matching}
 \acro{SIlog}{scale invariant logarithmic error}
 \acro{SRD}{squared relative distance}
 \acro{SSIM}{structural similarity}
 \acro{tMAE}{mean-absolute thresholded error}
 \acro{ToF}{time-of-flight}
 \acro{tRMSE}{root-mean-squared thresholded error}
\end{acronym}
\definecolor{dai_ligth_grey}{RGB}{230,230,230}
\definecolor{dai_ligth_grey20K}{RGB}{200,200,200}
\definecolor{dai_ligth_grey40K}{RGB}{158,158,158}
\definecolor{dai_ligth_grey60K}{RGB}{112,112,112}
\definecolor{dai_ligth_grey80K}{RGB}{68,68,68}

\definecolor{dai_petrol}{RGB}{0,103,127}
\definecolor{dai_petrol20K}{RGB}{0,86,106}
\definecolor{dai_petrol40K}{RGB}{0,67,85}
\definecolor{dai_petrol80}{RGB}{0,122,147}
\definecolor{dai_petrol60}{RGB}{80,151,171}
\definecolor{dai_petrol40}{RGB}{121,174,191}
\definecolor{dai_petrol20}{RGB}{166,202,216}

\definecolor{dai_deepred}{RGB}{113,24,12}
\definecolor{dai_deepred20K}{RGB}{90,19,10}
\definecolor{dai_deepred40K}{RGB}{68,14,7}

\definecolor{apfelgruen}{RGB}{140, 198, 62}
\definecolor{orange}{RGB}{244, 111, 33}
\definecolor{lila}{RGB}{128, 10, 145}
\definecolor{anthrazit}{RGB}{19, 31, 31}
\newcommand{\resultsPath}{fig/results}
\newcommand{\lidarColor}{black}
\newcommand{\psmColor}{dai_ligth_grey40K}
\newcommand{\sgmColor}{dai_petrol}
\newcommand{\sparseColor}{apfelgruen}
\newcommand{\monodepthColor}{dai_deepred}

\newcommand{\lidarLegend}{Lidar (int.)}
\newcommand{\psmLegend}{Deep Stereo \cite{Chang2018}}
\newcommand{\sgmLegend}{Trad. Stereo \cite{Hirschmuller2008}}
\newcommand{\sparseLegend}{RGB+Lidar \cite{ma2018sparse}}
\newcommand{\sparseLegendFull}{RGB+Lidar \cite{ma2018sparse}}
\newcommand{\monodepthLegend}{Monocular \cite{Godard2017}}
\newcommand{\monodepthLegendFt}{Monocular \cite{Godard2017}}

\newcommand{\topdownview}[2]{
\resizebox{#2}{!}{
\begin{tikzpicture}

\begin{axis}[
        axis on top,% ----
        width=5cm,
        scale only axis,
        enlargelimits=false,
        xmin=5,
        xmax=28,
        ymin=0,
        ymax=10,
        width=5.6cm,
        height=2cm,
        ytick={0,5,10},
        ]

      \addplot[thick] graphics[xmin=5,ymin=0,xmax=28,ymax=10] {#1};
    \end{axis}
\end{tikzpicture}
}
}

\newcommand{\DepthMetric}[4]{
\resizebox{\columnwidth}{!}{
\begin{tikzpicture}
				\begin{axis}
					[
					#3,
					xlabel=Distance {[m]},
					legend pos=north west,
					legend cell align=left,
					legend style={font=\footnotesize},
					legend columns=2,
					grid,
					xmin=10, xmax=26,
					]
					
				\addplot+ 	[
							very thick,
							solid, 
							\lidarColor,
							mark=none,
							]
				table[x index=0,y index=#2,col sep=space]{\resultsPath/lidar_interpolated/lidar_interpolated_#1_day_clear_metrics_binned.txt};
				\addlegendentry{\lidarLegend}
				
				\addplot+ 	[
							very thick,
							solid, 
							\psmColor,
							mark=none,
							]
				table[x index=0,y index=#2,col sep=space]{\resultsPath/psm/psm_#1_day_clear_metrics_binned.txt};
				\addlegendentry{\psmLegend}

				\addplot+ 	[
							very thick,
							solid, 
							\sgmColor,
							mark=none,
							]
				table[x index=0,y index=#2,col sep=space]{\resultsPath/sgm/sgm_#1_day_clear_metrics_binned.txt};
				\addlegendentry{\sgmLegend}
				
				\addplot+ 	[
							very thick,
							solid, 
							\sparseColor,
							mark=none,
							]
				table[x index=0,y index=#2,col sep=space]{\resultsPath/sparse2dense_full/sparse2dense_full_#1_day_clear_metrics_binned.txt};
				\addlegendentry{\sparseLegendFull}				
				
				\addplot+ 	[
							very thick,
							solid, 
							\monodepthColor,
							mark=none,
							]
				table[x index=0,y index=#2,col sep=space]{\resultsPath/monodepth_ft/monodepth_ft_#1_day_clear_metrics_binned.txt};
				\addlegendentry{\monodepthLegendFt}
				
				\addplot+ 	[
							very thick,
							densely dashed, 
							\lidarColor,
							mark=none,
							]
				table[x index=0,y index=#2,col sep=space]{\resultsPath/lidar_interpolated/lidar_interpolated_#1_night_clear_metrics_binned.txt};
				
				\addplot+ 	[
							very thick,
							densely dashed, 
							\psmColor,
							mark=none,
							]
				table[x index=0,y index=#2,col sep=space]{\resultsPath/psm/psm_#1_night_clear_metrics_binned.txt};

				\addplot+ 	[
							very thick,
							densely dashed, 
							\sgmColor,
							mark=none,
							]
				table[x index=0,y index=#2,col sep=space]{\resultsPath/sgm/sgm_#1_night_clear_metrics_binned.txt};
							
				\addplot+ 	[
							very thick,
							densely dashed, 
							\sparseColor,
							mark=none,
							]
				table[x index=0,y index=#2,col sep=space]{\resultsPath/sparse2dense_full/sparse2dense_full_#1_night_clear_metrics_binned.txt};

				\addplot+ 	[
							very thick,
							densely dashed,  
							\monodepthColor,
							mark=none,
							]
				table[x index=0,y index=#2,col sep=space]{\resultsPath/monodepth_ft/monodepth_ft_#1_night_clear_metrics_binned.txt};				

#4				
				\end{axis}
	\end{tikzpicture}
}
}

\newcommand{\MethodComparisonErrorMetric}[6]{
\begin{tikzpicture}
				\begin{axis}
					[
					#5,
					legend cell align=left,
					legend style={font=\footnotesize},
					legend columns=2,
					grid,
					]
					
				\addplot+ 	[
							very thick,
							solid, 
							\lidarColor,
							mark=none,
							]
				table[x index=0,y index=#1,col sep=comma]{\resultsPath/lidar_interpolated/lidar_interpolated_#2_day_#3_metrics#4.txt};
				\addlegendentry{\lidarLegend}
				
				\addplot+ 	[
							very thick,
							solid, 
							\psmColor,
							mark=none,
							]
				table[x index=0,y index=#1,col sep=comma]{\resultsPath/psm/psm_#2_day_#3_metrics#4.txt};
				\addlegendentry{\psmLegend}
				
				\addplot+ 	[
							very thick,
							solid, 
							\sgmColor,
							mark=none,
							]
				table[x index=0,y index=#1,col sep=comma]{\resultsPath/sgm/sgm_#2_day_#3_metrics#4.txt};
				\addlegendentry{\sgmLegend}
				
				\addplot+ 	[
							very thick,
							solid, 
							\sparseColor,
							mark=none,
							]
				table[x index=0,y index=#1,col sep=comma]{\resultsPath/sparse2dense_full/sparse2dense_full_#2_day_#3_metrics#4.txt};
				\addlegendentry{\sparseLegendFull}			

				\addplot+ 	[
							very thick,
							solid, 
							\monodepthColor,
							mark=none,
							]
				table[x index=0,y index=#1,col sep=comma]{\resultsPath/monodepth_ft/monodepth_ft_#2_day_#3_metrics#4.txt};
				\addlegendentry{\monodepthLegendFt}							
				
				\addplot+ 	[
							very thick,
							densely dashed, 
							\lidarColor,
							mark=none,
							]
				table[x index=0,y index=#1,col sep=comma]{\resultsPath/lidar_interpolated/lidar_interpolated_#2_night_#3_metrics#4.txt};
				
				\addplot+ 	[
							very thick,
							densely dashed, 
							\psmColor,
							mark=none,
							]
				table[x index=0,y index=#1,col sep=comma]{\resultsPath/psm/psm_#2_night_#3_metrics#4.txt};
				
				\addplot+ 	[
							very thick,
							densely dashed, 
							\sgmColor,
							mark=none,
							]
				table[x index=0,y index=#1,col sep=comma]{\resultsPath/sgm/sgm_#2_night_#3_metrics#4.txt};				

				\addplot+ 	[
							very thick,
							densely dashed, 
							\sparseColor,
							mark=none,
							]
				table[x index=0,y index=#1,col sep=comma]{\resultsPath/sparse2dense_full/sparse2dense_full_#2_night_#3_metrics#4.txt};
				
				\addplot+ 	[
							very thick,
							densely dashed, 
							\monodepthColor,
							mark=none,
							]
				table[x index=0,y index=#1,col sep=comma]{\resultsPath/monodepth_ft/monodepth_ft_#2_night_#3_metrics#4.txt};

#6
				\end{axis}
	\end{tikzpicture}
}

\newcommand{\DepthDistribution}[1]{
\resizebox{#1}{!}{
\begin{tikzpicture}

				\begin{axis}
					[
					height=0.57\linewidth,
					width=\linewidth,
					legend pos=north east,
					xlabel=Distance $d$ {[m]},
					ylabel=Number of points,
					legend cell align=left,
					legend style={font=\footnotesize},
					xmin=5, xmax=28,
					grid,
					yticklabel style={
					        /pgf/number format/fixed,
					},
					scaled y ticks=false,
					]
					
				\addplot+ 	[
							very thick,
							solid, 
							black,
							mark=none,
							]
				table[x index=0,y index=1,col sep=space]{\resultsPath/intermetric/intermetric_scene1_hist.txt};
				\addlegendentry{Pedestrian Zone}
				
				\addplot+ 	[
							very thick,
							solid, 
							dai_deepred,
							mark=none,
							]
				table[x index=0,y index=1,col sep=space]{\resultsPath/intermetric/intermetric_scene2_hist.txt};
				\addlegendentry{Residential Area}

				\addplot+ 	[
							very thick,
							solid, 
							dai_petrol,
							mark=none,
							]
				table[x index=0,y index=1,col sep=space]{\resultsPath/intermetric/intermetric_scene3_hist.txt};
				\addlegendentry{Construction Area}
				
				\addplot+ 	[
							very thick,
							solid, 
							apfelgruen,
							mark=none,
							]
				table[x index=0,y index=1,col sep=space]{\resultsPath/intermetric/intermetric_scene4_hist.txt};	
				\addlegendentry{Highway}			
				
				\end{axis}
	\end{tikzpicture}
}
}

\newcommand{\qualitativeImagesLine}[4]{
\includegraphics[width=0.145\textwidth]{\resultsPath/rgb/rgb_#4_#2_#3_#1_clahe} &
\includegraphics[width=0.145\textwidth]{\resultsPath/lidar_interpolated/lidar_interpolated_#4_#2_#3_#1} &
\includegraphics[width=0.145\textwidth]{\resultsPath/psm/psm_#4_#2_#3_#1} &
\includegraphics[width=0.145\textwidth]{\resultsPath/sgm/sgm_#4_#2_#3_#1} &
\includegraphics[width=0.145\textwidth]{\resultsPath/sparse2dense_full/sparse2dense_full_#4_#2_#3_#1} &
\includegraphics[width=0.145\textwidth]{\resultsPath/monodepth_ft/monodepth_ft_#4_#2_#3_#1}
}

\newcommand{\qualitativeImagesAll}[1]{
\setlength\tabcolsep{1pt}
\renewcommand{\arraystretch}{0.8}
\centering
\begin{tabular}{>{\centering\arraybackslash}m{0.5cm} >{\centering\arraybackslash}m{0.5cm} >{\centering\arraybackslash}m{0.2cm}  >{\centering\arraybackslash}m{0.15\textwidth} >{\centering\arraybackslash}m{0.15\textwidth} >{\centering\arraybackslash}m{0.15\textwidth} >{\centering\arraybackslash}m{0.15\textwidth} >{\centering\arraybackslash}m{0.15\textwidth} >{\centering\arraybackslash}m{0.15\textwidth}}

\multirow{8}{*}{\rotatebox[origin=l]{90}{\parbox[c]{10cm}{\centering \textbf{\textsc{Day}}}}} & & & RGB & \lidarLegend & \psmLegend & \sgmLegend & \sparseLegend & \monodepthLegend \\\cline{1-2}
& \vspace*{5pt}\rotatebox[origin=l]{90}{\textbf{clear}} & & \qualitativeImagesLine{0}{day}{clear}{#1} \\\cline{2-2}
& \multirow{5}{*}{\rotatebox[origin=l]{90}{\parbox[c]{4.3cm}{\centering \textbf{fog}}}} & \rotatebox[origin=l]{90}{\footnotesize{\unit[20]{m}}} & \qualitativeImagesLine{20}{day}{fog}{#1} \\
& & \rotatebox[origin=l]{90}{\footnotesize{\unit[30]{m}}} & \qualitativeImagesLine{30}{day}{fog}{#1} \\
& & \rotatebox[origin=l]{90}{\footnotesize{\unit[50]{m}}} & \qualitativeImagesLine{50}{day}{fog}{#1} \\
& & \rotatebox[origin=l]{90}{\footnotesize{\unit[70]{m}}} & \qualitativeImagesLine{70}{day}{fog}{#1} \\
& & \rotatebox[origin=l]{90}{\footnotesize{\unit[100]{m}}} & \qualitativeImagesLine{100}{day}{fog}{#1} \\\cline{2-2}
& \multirow{2}{*}{\rotatebox[origin=l]{90}{\parbox[c]{1.7cm}{\centering \textbf{rain}}}} & \rotatebox[origin=l]{90}{\footnotesize{\unit[15]{mm}}} & \qualitativeImagesLine{15}{day}{rain}{#1} \\
& & \rotatebox[origin=l]{90}{\footnotesize{\unit[55]{mm}}} & \qualitativeImagesLine{55}{day}{rain}{#1} \\\cline{1-2}
&&&&&&&&\\\cline{1-2}
\multirow{8}{*}{\rotatebox[origin=l]{90}{\parbox[c]{9cm}{\centering \textbf{\textsc{Night}}}}} & \vspace*{5pt}\rotatebox[origin=l]{90}{\textbf{clear}} & & \qualitativeImagesLine{0}{night}{clear}{#1} \\\cline{2-2}
& \multirow{5}{*}{\rotatebox[origin=l]{90}{\parbox[c]{4.3cm}{\centering \textbf{fog}}}} & \rotatebox[origin=l]{90}{\footnotesize{\unit[20]{m}}} & \qualitativeImagesLine{20}{night}{fog}{#1} \\
& & \rotatebox[origin=l]{90}{\footnotesize{\unit[30]{m}}} & \qualitativeImagesLine{30}{night}{fog}{#1} \\
& & \rotatebox[origin=l]{90}{\footnotesize{\unit[50]{m}}} & \qualitativeImagesLine{50}{night}{fog}{#1} \\
& & \rotatebox[origin=l]{90}{\footnotesize{\unit[70]{m}}} & \qualitativeImagesLine{70}{night}{fog}{#1} \\
& & \rotatebox[origin=l]{90}{\footnotesize{\unit[100]{m}}} & \qualitativeImagesLine{100}{night}{fog}{#1} \\\cline{2-2}
& \multirow{2}{*}{\rotatebox[origin=l]{90}{\parbox[c]{1.7cm}{\centering \textbf{rain}}}} & \rotatebox[origin=l]{90}{\footnotesize{\unit[15]{mm}}} & \qualitativeImagesLine{15}{night}{rain}{#1} \\
& & \rotatebox[origin=l]{90}{\footnotesize{\unit[55]{mm}}} & \qualitativeImagesLine{55}{night}{rain}{#1} \\\cline{1-2}
& &	& \multicolumn{6}{c}{\includegraphics[width=0.915\linewidth]{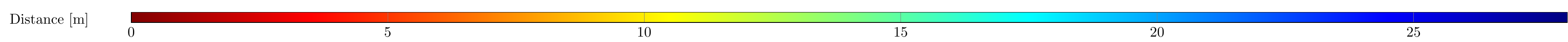}	} \\

\end{tabular}

}

\newcommand{\qualitativeResultsAll}[2]{
\setlength\tabcolsep{1pt}
\begin{tabular}{>{\centering\arraybackslash}m{0.5cm} >{\centering\arraybackslash}m{0.16\textwidth} >{\centering\arraybackslash}m{0.16\textwidth} >{\centering\arraybackslash}m{0.16\textwidth} >{\centering\arraybackslash}m{0.16\textwidth} >{\centering\arraybackslash}m{0.16\textwidth} >{\centering\arraybackslash}m{0.16\textwidth}}
	& Ground Truth & \lidarLegend & \psmLegend & \sgmLegend & \sparseLegendFull & \monodepthLegendFt \\
	\rotatebox[origin=l]{90}{\footnotesize{\textsc{Day}}} & \includegraphics[width=\linewidth]{\resultsPath/intermetric/intermetric_#1} &
	\includegraphics[width=\linewidth]{\resultsPath/lidar_interpolated/lidar_interpolated_#1_day_#2_0} &
	\includegraphics[width=\linewidth]{\resultsPath/psm/psm_#1_day_#2_0} &
	\includegraphics[width=\linewidth]{\resultsPath/sgm/sgm_#1_day_#2_0} &
	\includegraphics[width=\linewidth]{\resultsPath/sparse2dense_full/sparse2dense_full_#1_day_#2_0} &
	\includegraphics[width=\linewidth]{\resultsPath/monodepth_ft/monodepth_ft_#1_day_#2_0} 
	\\
	
	\rotatebox[origin=l]{90}{\footnotesize{\textsc{Night}}} & \includegraphics[width=\linewidth]{\resultsPath/intermetric/intermetric_#1} &
	\includegraphics[width=\linewidth]{\resultsPath/lidar_interpolated/lidar_interpolated_#1_night_#2_0} &
	\includegraphics[width=\linewidth]{\resultsPath/psm/psm_#1_night_#2_0} &
	\includegraphics[width=\linewidth]{\resultsPath/sgm/sgm_#1_night_#2_0} &
	\includegraphics[width=\linewidth]{\resultsPath/sparse2dense_full/sparse2dense_full_#1_night_#2_0} &
	\includegraphics[width=\linewidth]{\resultsPath/monodepth_ft/monodepth_ft_#1_night_#2_0}
	\\
			
	& \multicolumn{6}{c}{\includegraphics[width=0.98\linewidth]{colorbar_depth}	} \\
	
	\rotatebox[origin=l]{90}{\footnotesize{\textsc{Day}}} & \includegraphics[width=\linewidth]{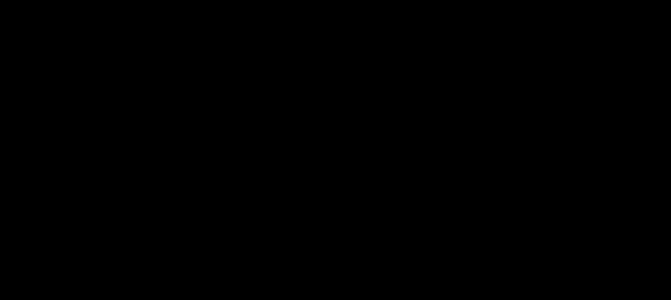} & 
	\includegraphics[width=\linewidth]{\resultsPath/lidar_interpolated/lidar_interpolated_#1_day_#2_error_image} &
	\includegraphics[width=\linewidth]{\resultsPath/psm/psm_#1_day_#2_error_image} &
	\includegraphics[width=\linewidth]{\resultsPath/sgm/sgm_#1_day_#2_error_image} &
	\includegraphics[width=\linewidth]{\resultsPath/sparse2dense_full/sparse2dense_full_#1_day_#2_error_image} &
	\includegraphics[width=\linewidth]{\resultsPath/monodepth_ft/monodepth_ft_#1_day_#2_error_image} 
	 \\ 
		
	\rotatebox[origin=l]{90}{\footnotesize{\textsc{Night}}}& \includegraphics[width=\linewidth]{\resultsPath/intermetric/intermetric_error} & 
	\includegraphics[width=\linewidth]{\resultsPath/lidar_interpolated/lidar_interpolated_#1_night_#2_error_image} &
	\includegraphics[width=\linewidth]{\resultsPath/psm/psm_#1_night_#2_error_image} &
	\includegraphics[width=\linewidth]{\resultsPath/sgm/sgm_#1_night_#2_error_image} &
	\includegraphics[width=\linewidth]{\resultsPath/sparse2dense_full/sparse2dense_full_#1_night_#2_error_image} &
	\includegraphics[width=\linewidth]{\resultsPath/monodepth_ft/monodepth_ft_#1_night_#2_error_image} 
	\\ 
			
	& \multicolumn{6}{c}{\includegraphics[width=0.98\linewidth]{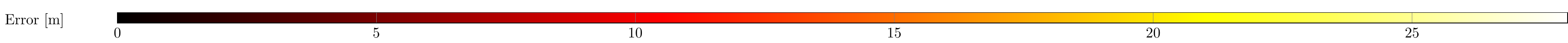}	} \\
		
	\rotatebox[origin=l]{90}{\footnotesize{\textsc{Day}}} & \topdownview{\resultsPath/intermetric/intermetric_#1_topview}{\linewidth} &
	\topdownview{\resultsPath/lidar_interpolated/lidar_interpolated_#1_day_#2_topview}{\linewidth} &
	\topdownview{\resultsPath/psm/psm_#1_day_#2_topview}{\linewidth} &
	\topdownview{\resultsPath/sgm/sgm_#1_day_#2_topview}{\linewidth} &
	\topdownview{\resultsPath/sparse2dense_full/sparse2dense_full_#1_day_#2_topview}{\linewidth} &
	\topdownview{\resultsPath/monodepth_ft/monodepth_ft_#1_day_#2_topview}{\linewidth}
	\\

	\rotatebox[origin=l]{90}{\footnotesize{\textsc{Night}}} & \topdownview{\resultsPath/intermetric/intermetric_#1_topview}{\linewidth} &
	\topdownview{\resultsPath/lidar_interpolated/lidar_interpolated_#1_night_#2_topview}{\linewidth} &
	\topdownview{\resultsPath/psm/psm_#1_night_#2_topview}{\linewidth} &
	\topdownview{\resultsPath/sgm/sgm_#1_night_#2_topview}{\linewidth} &
	\topdownview{\resultsPath/sparse2dense_full/sparse2dense_full_#1_night_#2_topview}{\linewidth} &
	\topdownview{\resultsPath/monodepth_ft/monodepth_ft_#1_night_#2_topview}{\linewidth} \\

\end{tabular}
}

\begin{document}

%%%%%%%%% TITLE
\title{Pixel-Accurate Depth Evaluation in Realistic Driving Scenarios}

\author{
	\hspace{-0.12in}Tobias Gruber$^{1,3}$\hspace{0.06in}
	Mario Bijelic$^{1,3}$\hspace{0.12in}
	Felix Heide$^{2,4}$\hspace{0.12in}
	Werner Ritter$^{1}$\hspace{0.12in}
	Klaus Dietmayer$^{3}$\hspace{0.12in} \vspace{3pt}
	\\ 
	\textsuperscript{1}Daimler AG\hspace{0.15in}
	\textsuperscript{2}Algolux\hspace{0.15in}
	\textsuperscript{3}Ulm University\hspace{0.15in}
	\textsuperscript{4}Princeton University\hspace{0.15in} 
}

\maketitle
\ifthreedvfinal\thispagestyle{empty}\fi

%%%%%%%%% ABSTRACT
\begin{abstract}
This work introduces an evaluation benchmark for depth estimation and completion using high-resolution depth measurements with angular resolution of up to \arcsec{25} (arcsecond), akin to a 50 megapixel camera with per-pixel depth available.
Existing datasets, such as the KITTI benchmark~\cite{geiger2013vision}, provide only sparse reference measurements with an order of magnitude lower angular resolution -- these sparse measurements are treated as ground truth by existing depth estimation methods.
We propose an evaluation methodology in four characteristic automotive scenarios recorded in varying weather conditions (day, night, fog, rain). As a result, our benchmark allows us to evaluate the robustness of depth sensing methods in adverse weather and different driving conditions. Using the proposed evaluation data, we demonstrate that current stereo approaches provide significantly more stable depth estimates than monocular methods and lidar completion in adverse weather. Data and code are available at \small{\url{https://github.com/gruberto/PixelAccurateDepthBenchmark.git}}.
\end{abstract}

%%%%%%%%% BODY TEXT

%------------------------------------------------------------------------
\section{Introduction}
3D scene understanding is one of the key challenges for safe autonomous driving, and the critical depth measurement and processing methods are a very active areas of research. 
Depth information can be captured using a variety of different sensing modalities, either passive or active.
Passive methods can be classified into stereo methods~\cite{Chang2018,Hirschmuller2008,Kendall2017,pilzer2018unsupervised}, which, inspired by the human visual system, extract depth from parallax in intensity images, \ac{SfM} \cite{koenderink1991affine,li2019learning,torr1999feature,Ummenhofer2017demon,Zhou2017}, and mono\-cular depth prediction \cite{Chen2018b,eigen2014depth,Godard2017,Laina2016,saxena2006learning}. Monocular depth estimation methods attempt to extract depth from cues such as defocus \cite{subbarao1994depth}, texture gradient and size perspective from a single image only. All of these passive sensing systems suffer in low light and at night, when the measured intensity is too low to robustly match image content.
Active sensing methods, such as lidar systems and \ac{ToF} cameras, overcome this challenge by relying on active illumination for depth measurements.
Specifically, lidar systems \cite{schwarz2010lidar} achieve large distances by focusing light into multiple beams which are mechanically scanned. This sequential acquisition fundamentally limits the angular resolution by the scanning mechanics, prohibiting semantic scene understanding at large distances where only a coarse sample distribution is available. As a result, a variety of algorithms for depth completion \cite{chen2018estimating,jaritz2018sparse,ma2018sparse} have been proposed recently.
Existing correlation \ac{ToF} cameras \cite{hansard2012time,kolb2010time,lange00tof} or structured light cameras \cite{achar2017epipolar,otoole2014temporal,o2012primal,scharstein2003high}, provide accurate high-resolution depth for close ranges indoors but suffer at long ranges in outdoor scenes due to strong ambient illumination and modulation frequency limitations.

\begin{figure}
\centering
\resizebox{\columnwidth}{!}{
\setlength\tabcolsep{0pt}
\begin{tabular}{cm{0.8cm}}
\setlength\tabcolsep{1.5pt}
\begin{tabular}{cc}
	\includegraphics[width=0.5\columnwidth]{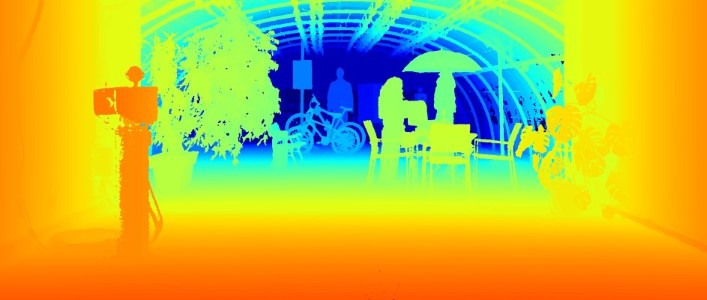} &
	\includegraphics[width=0.5\columnwidth]{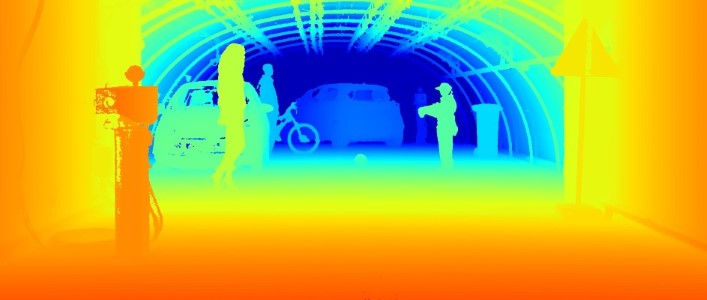} \\
	\includegraphics[width=0.5\columnwidth]{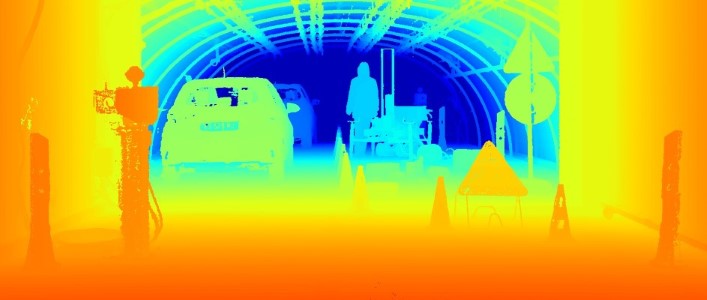} &
	\includegraphics[width=0.5\columnwidth]{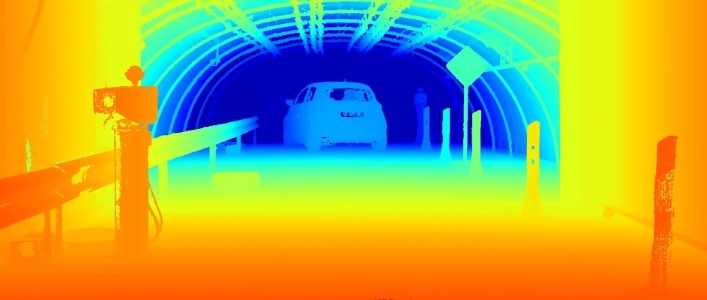} \\	\end{tabular} &
	\includegraphics[height=0.441\columnwidth]{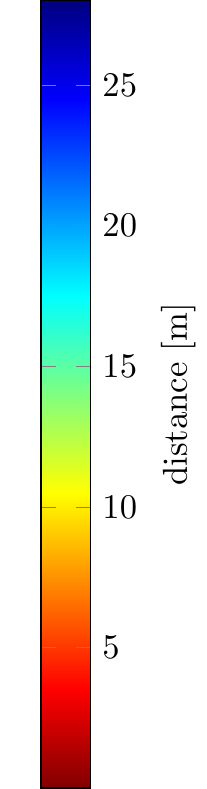} \\
\end{tabular}
	}
	\vspace*{-8pt}
	\caption{We propose a depth evaluation framework using ground truth depth data acquired at an angular resolution of up to \arcsec{25} (arcsecond). We evaluate depth sensing under realistic weather situations and different automotive scenarios, including a pedestrian zone, residential area, construction, area and highway.}
	\label{fig:groundtruth_examples}
	\vspace*{-15pt}
\end{figure}

In order to evaluate the performance of such diverse sensing approaches for autonomous driving, empirical datasets under realistic road conditions and meaningful metrics are required. Captured with previous generation hardware, the 
NYUdepth v2 \cite{silberman2012indoor} and KITTI \cite{geiger2013vision} datasets are established for developing and evaluating depth estimation methods.
While NYUdepth v2 contains a large variety of close-range indoor scenes captured by a Microsoft Kinect RGB-D camera \cite{zhang2012microsoft}, the KITTI dataset provides real-world street view scenarios recorded with a stereo camera and a lidar sensor.
Although the sparse depth measurements of the KITTI dataset have been accumulated over multiple laser scans \cite{uhrig2017sparsity}, the resulting ``ground truth'' depth provides only \arcsec{300} horizontal and \arcsec{700} vertical angular sampling. To detect the legs of a pedestrian at \unit[150]{m} distance an angular resolution of less than \arcsec{25} would be required.
In this work, we present an evaluation framework with pixel-wise annotated ground truth depth, as visualized in Figure~\ref{fig:groundtruth_examples}, at an angular resolution of \arcsec{25}. This resolution corresponds to a 50 mega-pixel camera (similar to KITTI \cite{geiger2013vision}: \unit[4.65]{\textmu m} pitch, \unit[4.4]{mm} focal length)  which enables semantic understanding tasks at large distances, such as pedestrian detection. We also record and provide stereo and lidar measurements acquired with recent state-of-the-art automotive sensors.

The robustness of the sensing and processing chain is critical for autonomous driving, which mandates reliability in adverse weather situations such as rain, fog and snow.
Note that the evaluation in these corner cases is particularly challenging because lidar sensors \emph{fail in severe adverse weather}. For example, in dense fog or snow, the first peak lidar measurements are unusable due to severe back-scatter in these scenarios. As a result, existing driving datasets~\cite{geiger2013vision,huang2018apolloscape,xu2017end} do not cover these severe conditions. To acquire accurate depth in challenging weather situations, and with reproducible scenes and conditions, we record sensor data with the proposed sensor setup in a weather chamber \cite{colomb2004innovative} that provides reproducible fog and rain. Using pixel-wise ground truth depth from clear conditions, the proposed approach allows us to accurately evaluate the performance under varying fog visibilities and rain intensities.

Specifically, we make the following contributions: 
\begin{itemize}
	\item We introduce an automotive depth evaluation dataset (1,600 samples) in adverse weather conditions with high-resolution annotated ground truth of angular resolution \arcsec{25} -- an order of magnitude higher than existing lidar datasets with angular resolution of \arcsec{300}.
	\item We evaluate the performance state-of-the-art algorithms for stereo depth, depth from mono, lidar and sparse depth completion, in reproducible, finely adjusted adverse weather situations.
	\item We demonstrate that stereo vision performs significantly more stable in fog than lidar and monocular depth estimation approaches.
\end{itemize}

%------------------------------------------------------------------------
\section{Related Work}
%-------------------------------------------------------------------------

\paragraph{Depth Estimation and Depth Completion}

Depth estimation algorithms can be categorized based on their input data. While \ac{SfM} approaches \cite{koenderink1991affine,li2019learning,torr1999feature,Ummenhofer2017demon,Zhou2017} rely on sequentially captured image data, multi-view depth estimation \cite{Chang2018,Hirschmuller2008,Kendall2017,pilzer2018unsupervised} uses at least two different views of a scene which are simultaneously captured. Monocular depth estimation methods \cite{Chen2018b,eigen2014depth,Godard2017,Laina2016,saxena2006learning} tackle the severely ill-posed depth reconstruction by monocular cues such as texture variation, gradients, object size, defocus, color or haze. Over the last years, \acp{CNN}s have been shown to be well-suited for both \ac{SfM}, multi-view and monocular approaches, and they can be trained supervised by large RGB-D indoor datasets acquired by consumer \ac{ToF} cameras. Since the acquisition of ground truth depth in large outdoor environments is challenging, semi-supervised \cite{Kuznietsov2017} or even self-supervised approaches \cite{aleotti2018generative,garg2016unsupervised,Godard2017,pillai2018superdepth} have been proposed, tackling this challenge by solving proxy tasks such as stereo matching. Another body of work focuses on completing sparse lidar point clouds. Existing depth completion methods rely on contextual information from RGB images to obtain high-resolution depth \cite{chen2018estimating,jaritz2018sparse,ma2018sparse}.

%-------------------------------------------------------------------------
\vspace{-9pt}
\paragraph{Depth Datasets}

The development and evaluation of depth estimation algorithms require a large amount of representative data, particularly for learned estimation methods.
Scharstein and Szeliski~\cite{scharstein2002taxonomy} provided the Middlebury data set as an early testing environment for quantitative evaluation of stereo algorithms, where the ground truth was obtained by structured light.
While the Middlebury dataset contained, in the first version, only two samples, subsequent datasets, such as Make 3D \cite{saxena2009make3d} provided around 500 samples with ground truth measured by a custom-build 3D scanner, though with lower resolution of 55x305.
Facilitated by consumer depth cameras such as the Microsoft Kinect \cite{zhang2012microsoft}, a number of depth data sets for indoor scenes have been proposed \cite{dai2017scannet,shotton2013scene,silberman2011indoor,silberman2012indoor,song2015sun}.
In particular, the NYUdepth v2 data set \cite{silberman2012indoor} is a widely used dataset with around 1500 samples.
However, due to the limitations of consumer depth cameras in severe ambient light and modulation frequency limitations, these data sets only include indoor scenarios with limited ranges.
The KITTI Stereo 2015 benchmark \cite{menze2015object} has introduced 400 images of street scenes with ground truth depth acquired by a lidar system. In order to mitigate the sparsity of raw ground truth depth (4\,\% coverage), 7 laser scans are registered and accumulated, and moving objects are replaced with geometrically accurate 3D models leading to 19\,\% coverage.
In order to obtain sufficient data for learning algorithms, Uhrig et. al~\cite{uhrig2017sparsity} presented a method to generate denser depth maps (16\,\% coverage) by automatically accumulating 11 laser scans and removing outliers by stereo comparison. Note that, even with accumulation, the resulting depth maps are still not providing depth at the resolution of the image sensor~\cite{menze2015object}.
Synthetic data sets such as the New Tsukuba Stereo Data set \cite{martull2012realistic}, Virtual KITTI \cite{gaidon2016virtual} and Synthia \cite{ros2016synthia} offer the possibility to create a theoretically unlimited amount of data with dense and accurate ground truth depth. 
While these data sets are extremely valuable for pretraining algorithms \cite{richter2016playing}, they cannot replace real recordings for performance evaluation of real-world applications due to the synthetic-to-real domain gap \cite{peng2018syn2real}.
In this work, we aim to close the gap between range-limited indoor and ground-truth-limited outdoor scenarios.

%-------------------------------------------------------------------------
\vspace{-9pt}
\paragraph{Robust Perception in Adverse Weather}

Robust environment perception is critical for enabling autonomous driving (without remote operators) over the world and in all environmental conditions. To this end, it is critical that self-driving cars should not stop working when being faced with unknown sensor distortions and situations that have not been in the training distribution. To characterize sensor distortions, previous methods have been focused on evaluating automotive sensors in challenging situations such as dust, smoke, rain, fog and snow \cite{peynot2009towards}.
For these evaluations, testing facilities such as Cerema \cite{colomb2004innovative} and Carissma \cite{hasirlioglu2018challenges} provide reproducible adverse weather situations with defined and adjustable severity.
In particular, cameras suffer from reduced contrast because particles in the air cause scattering \cite{Bijelic2018,oakley1998improving}. 
Recently, the Robust Vision Challenge \cite{geiger2018robust} promotes 
the development of robust algorithms benchmarked on a variety of datasets.
However, the generalization and robustness to real adverse weather is significantly more difficult than dataset generalization because challenging weather conditions occur rarely and change quickly \cite{van2010roles}.
As a result, recent approaches evaluate robustness using synthetically extended datasets such as Foggy Cityscapes \cite{sakaridis2018semantic}.
In this work, we depart from these synthetic datasets and instead propose a non-synthetic, but reproducible, benchmark for depth estimation in adverse weather conditions such as rain and fog under controlled conditions.

%------------------------------------------------------------------------
\section{Sensor Setup and Calibration}\label{sec:sensor_setup}
\begin{figure}[t]
	\centering
	\includegraphics[width=\linewidth]{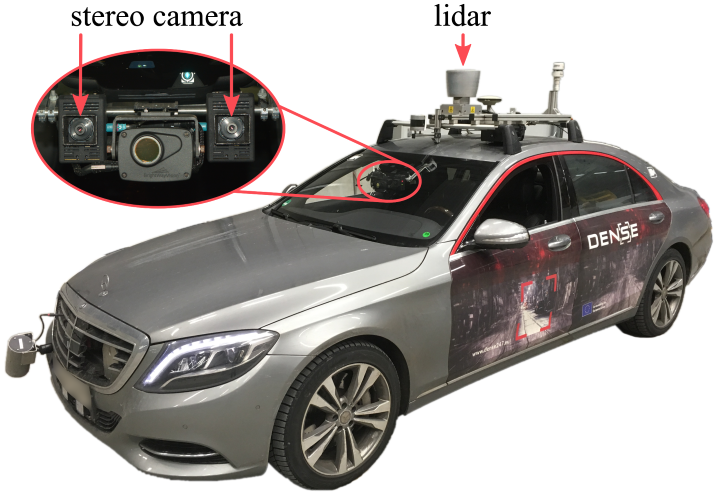}
	\vspace*{-20pt}
	\caption{Test vehicle equipped with a sensor suite including an Aptina AR0230 stereo camera and a Velodyne HDL64-S3 lidar system.}
	\label{fig:sensor_calibration}
	\vspace*{-15pt}
\end{figure}

To acquire realistic automotive sensor data, which serves as input for the depth evaluation methods assessed in this work, we equipped a research vehicle with a RGB stereo camera (Aptina AR0230, 1920x1024, 12bit)
and a lidar (Velodyne HDL64-S3, \unit[905]{nm}), see Figure~\ref{fig:sensor_calibration}. All sensors run in a \ac{ROS} environment and are time-synchronized by a \ac{PPS} signal provided by a proprietary \ac{IMU}. For the Velodyne lidar, both last and strongest return are recorded. Next, we describe the acquisition of the ground truth depth dataset that we use to evaluate the depth estimates obtained from the automotive sensor suite.

\vspace{-9pt}
\paragraph{Ground Truth Acquisition}

We acquire ground truth depth for static scenarios using a Leica ScanStation P30 laser scanner (360\textdegree/290\textdegree\, FOV, \unit[1550]{nm}, with up to 1M points per second, up to \arcsec{8} angular accuracy, and \unit[1.2]{mm} + \unit[10]{\ac{ppm}} range accuracy). 
To mitigate the effects of occlusions and to further increase the resolution, we accumulate multiple pointclouds at different overlapping positions.
At least three white sphere lidar targets with a diameter of \unit[145]{mm} at defined positions have to be detected for registration of the raw pointclouds. 
Each raw scan lasts about \unit[5]{min}, which limits the proposed high-resolution acquision approach to static scenarios.
Using the known positions of the targets in every raw pointcloud, transformations between these raw pointclouds are obtained by solving a linear least-squares problem.
As a result, we obtain a dense pointcloud ($\approx$ 50M points) with a uniformly distributed \emph{mean distance between neighboring points of only \unit[3]{mm}}, corresponding to an angular resolution of \arcsec{25}. We use the middle of the rear axis, identified by measuring the positions of the wheel hubs, as the origin of the target point clouds.
All sensors of the automotive sensor suite from Figure~\ref{fig:sensor_calibration} have been calibrated with respect to this ground truth pointcloud, which we describe in the following.

\vspace{-9pt}
\paragraph{Camera Calibration}

The intrinsic calibration of the stereo cameras is performed by detecting checkerboards with predefined field size \cite{zhang2000flexible}.
We recorded these checkerboards at different distances and viewpoints in order to obtain the camera matrix and distortion coefficients. 
In order to register the very dense ground truth point clouds to the camera coordinate systems (extrinsic calibration), multiple black-white targets are placed at known 3D positions.
By labeling the target positions in the images, an extrinsic calibration is obtained by solving the perspective-n-point problem using Levenberg-Marquardt non-linear least-squares optimization \cite{levenberg1944method,marquardt1963algorithm}.

\vspace{-9pt}
\paragraph{Lidar Registration}

While the resolution of scans from the Leica laser scanner facilitates the detection of the white sphere targets, it is challenging to detect these targets at larger distances in the Velodyne laser scan. 
Therefore, these lidar targets cannot be used for registration of both Leica and Velodyne pointclouds.
We use generalized \ac{ICP} \cite{segal2009generalized} for registration by minimizing the difference between two pointclouds, with the initial iterate shifted to the manually measured mounting position.

%------------------------------------------------------------------------
\section{Adverse Weather Dataset}
\begin{figure}[t]
\centering
\setlength\tabcolsep{1.5pt}
\begin{tabular}{cc}
\small{Pedestrian Zone} & \small{Residential Area} \\
\includegraphics[width=0.5\columnwidth]{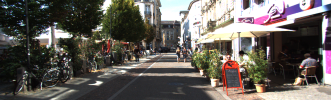} & 
\includegraphics[width=0.5\columnwidth]{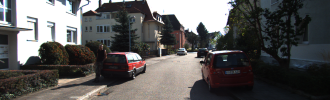} \\
\small{Construction Area} & \small{Highway} \\
\includegraphics[width=0.5\columnwidth]{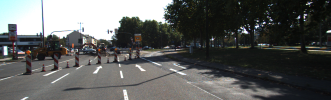} &
\includegraphics[width=0.5\columnwidth]{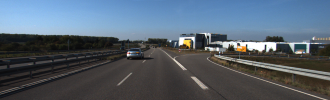}
\end{tabular}
\vspace*{-10pt}
\caption{Representative automotive scenarios, see \cite{geiger2013vision}, covered in this evaluation benchmark.}
\label{fig:kitti_examples}
\vspace*{-5pt}
\end{figure}

\begin{figure}[t]
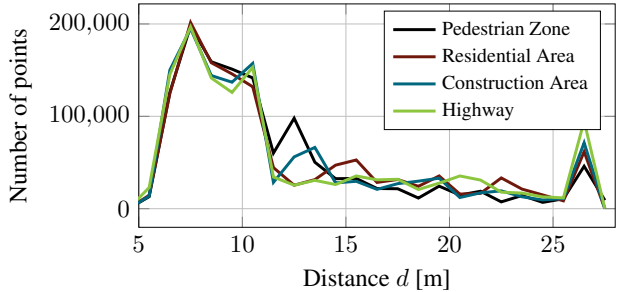

\DepthDistribution{\columnwidth}
\vspace*{-18pt}
\caption{The depth distribution of the proposed scenarios is primarily driven by the camera frustum. The scenario itself changes the distribution only slightly.}
\label{fig:depth_distribution}
\vspace*{-15pt}
\end{figure}

\begin{figure*}[t]
\centering
\setlength\tabcolsep{1.5pt}
\begin{tabular}{cccc}
	Pedestrian Zone & Residential Area & Construction Area & Highway \\
	\includegraphics[width=0.24\linewidth]{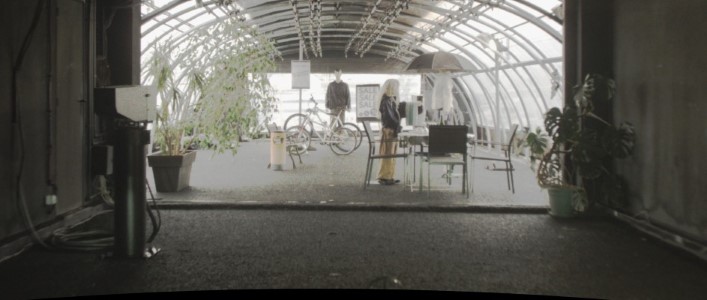} &
	\includegraphics[width=0.24\linewidth]{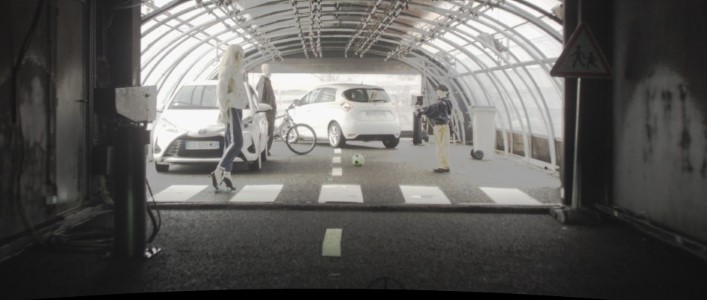} &
	\includegraphics[width=0.24\linewidth]{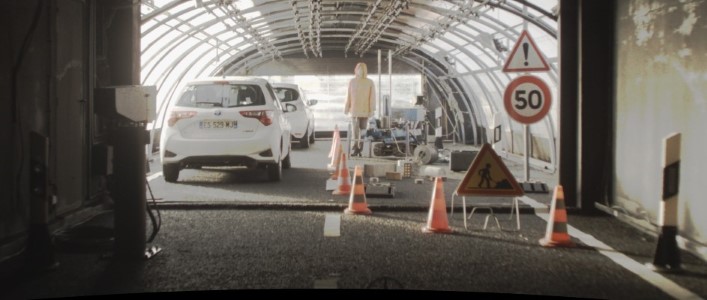} &
	\includegraphics[width=0.24\linewidth]{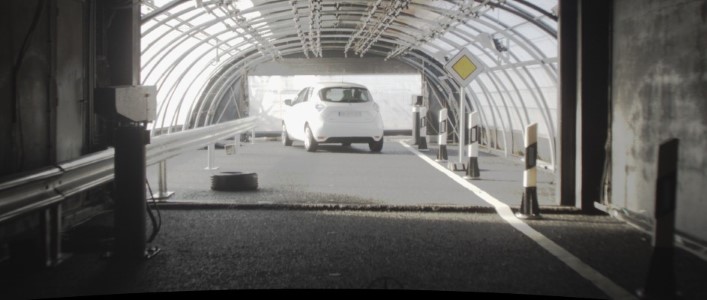} \\
	\includegraphics[width=0.24\linewidth]{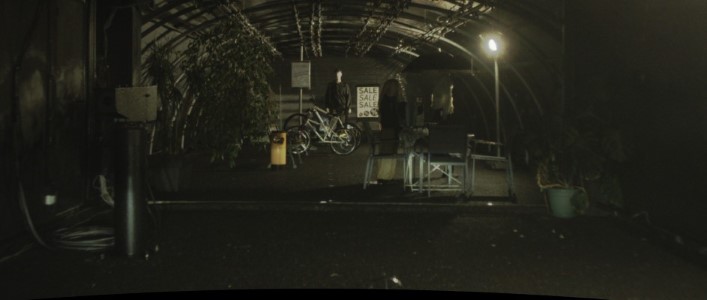} &
	\includegraphics[width=0.24\linewidth]{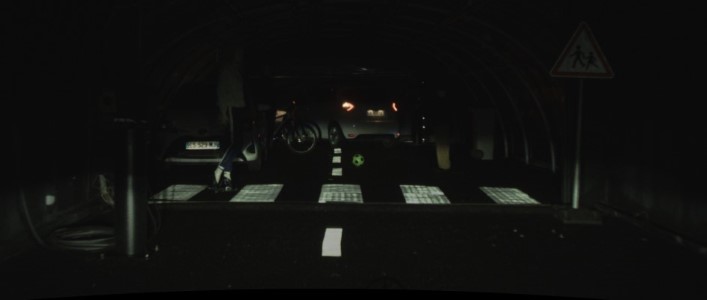} &
	\includegraphics[width=0.24\linewidth]{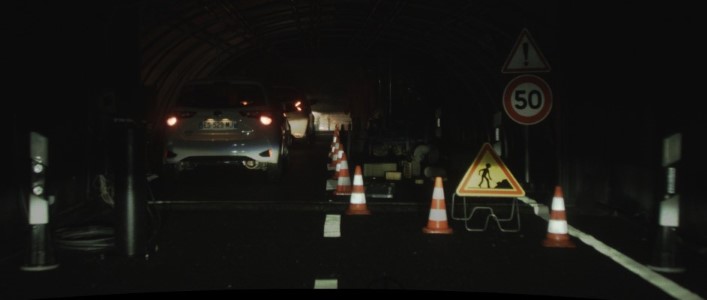} &
	\includegraphics[width=0.24\linewidth]{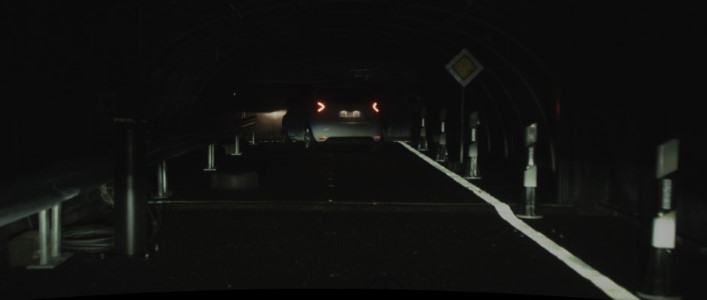} \\
	\includegraphics[width=0.24\linewidth]{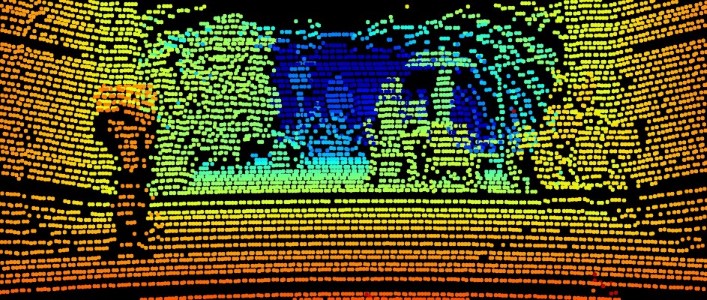} &
	\includegraphics[width=0.24\linewidth]{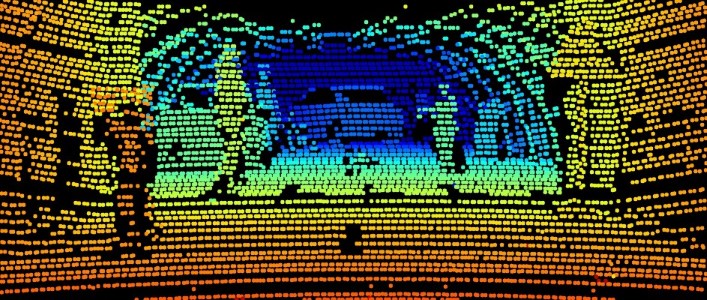} &
	\includegraphics[width=0.24\linewidth]{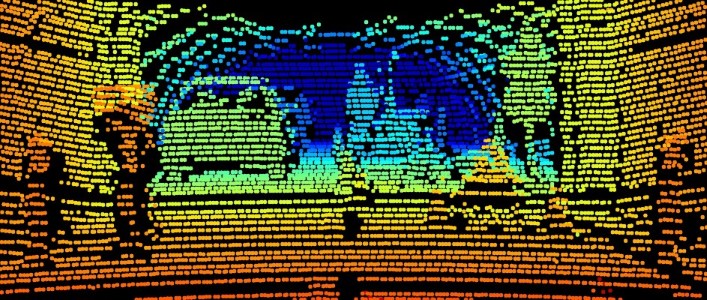} &
	\includegraphics[width=0.24\linewidth]{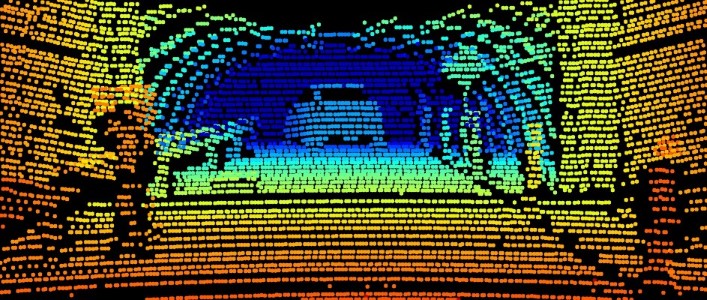} \\
	\includegraphics[width=0.24\linewidth]{\resultsPath/intermetric/intermetric_scene1} &
	\includegraphics[width=0.24\linewidth]{\resultsPath/intermetric/intermetric_scene2} &
	\includegraphics[width=0.24\linewidth]{\resultsPath/intermetric/intermetric_scene3} &
	\includegraphics[width=0.24\linewidth]{\resultsPath/intermetric/intermetric_scene4} \\
	\topdownview{\resultsPath/intermetric/intermetric_scene1_topview}{0.24\linewidth} &
	\topdownview{\resultsPath/intermetric/intermetric_scene2_topview}{0.24\linewidth} &
	\topdownview{\resultsPath/intermetric/intermetric_scene3_topview}{0.24\linewidth} &
	\topdownview{\resultsPath/intermetric/intermetric_scene4_topview}{0.24\linewidth} \\
	\end{tabular}
	\vspace*{-10pt}
	\caption{The proposed benchmark covers four different scenarios at day (first line) and night (second line). We provide state-of-the-art stereo camera and lidar sensors (third line) together with pixel-wise annotated ground truth (forth line). The last line shows a top-down view of each scenario. The large sensor on the left belongs to the visibility measurement system. The depth color coding is the same as in Figure~\ref{fig:groundtruth_examples}.}
	\label{fig:fogchamber_scenarios}
	\vspace*{-15pt}
\end{figure*}

We model the typical automotive outdoor scenarios from \cite{geiger2013vision}. Specifically, we setup the following four realistic scenarios: pedestrian zone, residential area, construction area and highway.
Real world examples of these scenarios are shown in Figure~\ref{fig:kitti_examples}, while our setup scenes are shown in Figure~\ref{fig:fogchamber_scenarios}.
We used mannequins as pedestrians in order to fix the scene during all measurements. 
These scenarios are recorded under different controlled weather and illumination conditions in a weather chamber \cite{colomb2004innovative}. 
The visible back part of the weather chamber, see Figure~\ref{fig:fogchamber_scenarios}, is constructed as greenhouse that is either transparent or covered by a black tarp and allows to achieve realistic daytime and night conditions.
After acquiring reference measurements in clear conditions, the whole chamber is flooded with fog. The fog density is tracked by the meteorological visibility $V$ defined by $V=-\ln\left(0.05\right)/\beta$, where $\beta$ is the atmospheric attenuation. 
As the fog slowly dissipates, visibility increases and the recordings are stopped after reaching $V=\unit[100]{m}$.
In order to obtain a larger number of samples, three dissipation runs have been performed.
For measurements in rain, two particular intensities at \unit[15 and 55]{mm/h/$\text{m}^2$} represent light and heavy rain.
In total, this benchmark consists of 10 randomly selected samples of each scenario (day/night) in clear, light rain, heavy rain and 17 visibility levels in fog (\unit[20-100]{m} in \unit[5]{m} steps), resulting in 1600 samples in total.

%------------------------------------------------------------------------
\section{Benchmark}\label{sec:proposed_benchmark}
In this section, we introduce all quantitative metrics and visualization methods used as evaluation methods in the remainder of this paper.
The evaluation is performed on 2.5D depth images because this domain allows for an immediate and intuitive comparison with the ground truth obtained by a depth camera or projected lidar points.

\vspace{-9pt}
\paragraph{Metrics}

We adopt established metrics for benchmarking depth estimation algorithms. 
Specifically, we use the metrics from the KITTI benchmark \cite{uhrig2017sparsity}, that is \ac{SIlog} \cite{eigen2014depth}, \ac{SRD}, \ac{ARD} and \ac{RMSE}. 
In addition, we evaluate \ac{MAE} and the threshold metric $\delta_i < 1.25^i$ for $i \in \left\{1,2,3 \right\}$.
Recently, Imran et al. \cite{imran2019depth} proposed \ac{tRMSE} and \ac{tMAE} as variants of the established \ac{RMSE} and \ac{MAE} metrics which we also add to our evaluation framework.
Moreover, as this work offers dense depth ground truth, we also assess depth map accuracy using dense image metrics, such as \ac{SSIM} and \ac{PSNR}. 
Since \ac{PSNR} is based on absolute distances, we introduce a variant using relative depth \ac{rPSNR}, see supplemental document. We provide an in-depth description and formal definition of each metric in the supplemental document.

\vspace{-9pt}
\paragraph{Binned Metrics}
The depth in a depth map is typically not uniformly distributed, as shown in Figure~\ref{fig:depth_distribution}.
By calculating the mean error of a depth image, errors at shorter distances contribute more to the mean than errors at larger distances. 
For a fair comparison of algorithms, we also provide binned evaluations where metrics are calculated in bins of approximately \unit[2]{m} and the mean of the bins gives the final result.
This ensures that every distance contributes equally to the evaluation metric.
\vspace{-9pt}
\paragraph{Top-Down View}

As additional qualitative visualization, we provide top-down views generated by projecting 2.5D depth images into 3D. 
However, projecting all points to the ground plane does not provide any meaningful top-down view because in such a top-view, points from the ground, from the ceiling, and from objects in between cannot be distinguished.
For improved visualization of the top-down view, we discretize the x-y plane into \unit[10x10]{cm} grid cells and squeeze the height by counting the number of points in each grid cell \cite{ku2018joint, yang2018pixor}.
As the number of points in a cell is decreasing with distance (see Figure~\ref{fig:depth_distribution}), we normalize the number of points according to their distance. 

%------------------------------------------------------------------------
\section{Evaluation}
\paragraph{Evaluated Methods}

A large body of work on depth estimation and depth completion has emerged over the recent years.
For brevity, we focus in this benchmark on one algorithm per algorithm category. Specifically, we compare \emph{Monodepth} \cite{Godard2017} as a representative method for monocular depth estimation, \ac{SGM} \cite{Hirschmuller2008} as traditional stereo and \emph{PSMnet} \cite{Chang2018} as deep stereo algorithms, and \emph{Sparse2Dense} \cite{ma2018sparse} as a depth completion method for lidar measurements using RGB image data. 
We set all algorithms up to estimate full-resolution depth maps, except for \emph{Monodepth} \cite{Godard2017} where we observed a substantial drop in performance and therefore resized the images to the native size the model was trained on.
Additionally, we finetuned the model on the training split of \cite{gruber2019gated2depth} where the same sensor setup has been utilized.
For \emph{Sparse2Dense} \cite{ma2018sparse}, we trained the network with 8000 points on KITTI in order to apply it to our projected lidar points.
Similar to the KITTI Depth Prediction benchmark \cite{uhrig2017sparsity}, we interpolate the results of methods that provide less than 100\,\% density with nearest neighbor interpolation.
Additionally, we crop the images for 270 pixels from the top, 170 pixels from the left, 20 pixels from the right and 20 pixels from the bottom to avoid boundary artifacts, e.g. missing lidar points in the top part of the image.

\vspace{-9pt}
\paragraph{Public Benchmark}

We will make all sensor data and the high-resolution ground truth data publicly available. 
All code for calculating the error metrics and for generating qualitative results (color-coded depth map, error map and top-down view) will be provided.
The dataset is high-resolution, and enables fine-grained evaluation in controlled adverse weather conditions.

%-------------------------------------------------------------------------
\vspace{-9pt}
\paragraph{Clear Weather Evaluation}

\begin{figure*}
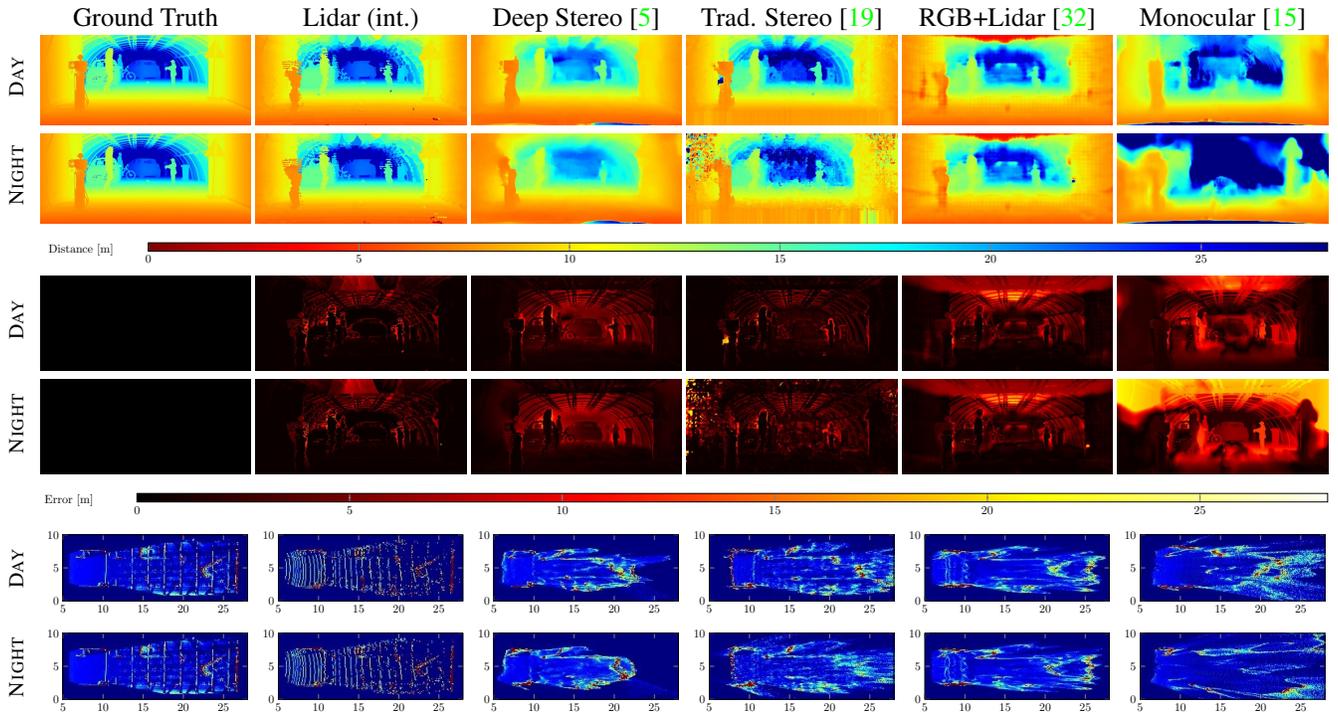

\centering
\vspace*{-10pt}
\qualitativeResultsAll{scene2}{clear}
\vspace*{-10pt}
\caption{Qualitative evaluation results of the residential area scenario at day and night comparison in clear conditions. For each benchmarked algorithm, we include the color-coded depth map (top), error map (middle) and top-down view (bottom).}
\label{fig:qualitative_results_clear}
\vspace*{-5pt}
\end{figure*}

\begin{table*}[t]
	\centering
	\resizebox{.99\linewidth}{!}{
	
\begin{tabular}{@{}>{\centering\arraybackslash}m{0.2cm} >{\centering\arraybackslash}m{0.2cm} lcccccccccccccc@{}}
		\toprule
		&&& \multicolumn{4}{|c|}{absolute} & \multicolumn{3}{|c|}{relative} & scale-invariant & \multicolumn{3}{|c|}{} & \multicolumn{3}{|c|}{full-depth} \\
		&&\textbf{Method}  & 
		\textbf{RMSE} & \textbf{tRMSE} & \textbf{MAE} & \textbf{tMAE} &
		\textbf{logRMSE} & \textbf{SRD} & \textbf{ARD} &
		\textbf{SIlog} &  
		$\boldsymbol{\delta_1}$ & $\boldsymbol{\delta_2}$ & $\boldsymbol{\delta_3}$  & 
		\textbf{SSIM} & \textbf{PSNR} & \textbf{rPSNR} \\
		
		&&& $\left[ \unit{m} \right]$  & $\left[ \unit{m} \right]$ & $\left[ \unit{m} \right]$ & $\left[ \unit{m} \right]$ &
		& & $\left[ \% \right]$ &
	    $100\log\left(\unit{m}\right)$ &
		$\left[ \% \right]$ & $\left[ \% \right]$ & $\left[ \% \right]$ &
		$\left[0-1\right]$ & $\left[ \unit{dB} \right]$ & $\left[ \unit{dB} \right]$ \\
		
		&&& $\downarrow$  & $\downarrow$ & $\downarrow$ & $\downarrow$ &
		$\downarrow$ & $\downarrow$ & $\downarrow$ &
		$\downarrow$ &
		$\uparrow$ & $\uparrow$ & $\uparrow$ &
		$\uparrow$ & $\downarrow$ & $\downarrow$ \\
		\midrule
		\multirow{10}{*}{\rotatebox[origin=l]{90}{\parbox[c]{4.5cm}{\centering \textbf{\textsc{Day}}}}}
		& \multirow{5}{*}{\rotatebox[origin=l]{90}{\parbox[c]{2cm}{\centering \textbf{\small{not binned}}}}}
		&\lidarLegend & 1.89 & \textbf{1.36} & \textbf{0.70} & \textbf{0.59} & \textbf{0.13} & \textbf{0.23} & \textbf{4.79} & \textbf{12.60} & \textbf{93.62} & 98.13 & 99.36 & 0.49 & 19.67 & 15.05  \\ 

		&&\psmLegend & 2.75 & 1.96 & 1.44 & 1.22 & 0.18 & 0.56 & 9.91 & 16.07 & 89.14 & 97.21 & 98.80 & \textbf{0.64} & 17.93 & 17.53  \\ 

		&&\sgmLegend & \textbf{1.90} & 1.40 & 0.96 & 0.86 & 0.14 & 0.27 & 8.12 & 13.32 & 90.74 & \textbf{98.44} & \textbf{99.50} & 0.47 & 20.82 & 16.43  \\ 

		&&\sparseLegendFull & 3.05 & 2.04 & 1.61 & 1.29 & 0.26 & 0.53 & 10.85 & 24.01 & 84.69 & 94.77 & 97.05 & 0.46 & 16.39 & \textbf{11.46}  \\ 

		&&\monodepthLegendFt & 5.01 & 2.67 & 2.93 & 2.01 & 0.33 & 2.78 & 27.12 & 29.12 & 73.73 & 88.87 & 95.10 & 0.47 & \textbf{14.45} & 18.96  \\ 

		\cmidrule{2-17}
		& \multirow{5}{*}{\rotatebox[origin=l]{90}{\parbox[c]{2cm}{\centering \textbf{\small{binned}}}}}
		&\lidarLegend & \textbf{2.41} & \textbf{1.91} & \textbf{1.51} & \textbf{1.33} & \textbf{0.23} & \textbf{1.07} & \textbf{24.24} & 14.85 & \textbf{82.03} & 89.10 & \textbf{91.46} & \textbf{0.54} & 15.08 & 4.40  \\ 

		&&\psmLegend & 3.82 & 2.63 & 3.10 & 2.38 & 0.35 & 4.18 & 50.56 & 15.66 & 69.50 & 84.12 & 88.01 & 0.52 & 11.92 & 1.10  \\ 

		&&\sgmLegend & 2.52 & 2.01 & 1.83 & 1.60 & 0.26 & 1.92 & 36.31 & \textbf{11.81} & 76.11 & \textbf{89.21} & 91.16 & 0.41 & 14.90 & 4.32  \\ 

		&&\sparseLegendFull & 3.82 & 2.61 & 2.89 & 2.24 & 0.39 & 1.75 & 33.07 & 24.52 & 66.30 & 82.63 & 86.36 & 0.44 & 11.73 & 0.92  \\ 

		&&\monodepthLegendFt & 7.49 & 3.42 & 6.33 & 3.07 & 0.54 & 31.62 & 143.44 & 26.30 & 56.63 & 75.92 & 82.59 & 0.36 & \textbf{8.15} & \textbf{-2.24} \\ 

		\midrule
		\multirow{10}{*}{\rotatebox[origin=l]{90}{\parbox[c]{4.5cm}{\centering \textbf{\textsc{Night}}}}}
		& \multirow{5}{*}{\rotatebox[origin=l]{90}{\parbox[c]{2cm}{\centering \textbf{\small{not binned}}}}}
		&\lidarLegend & \textbf{1.90} & \textbf{1.36} & \textbf{0.70} & \textbf{0.58} & \textbf{0.13} & \textbf{0.23} & \textbf{4.83} & \textbf{12.77} & \textbf{93.57} & \textbf{98.03} & \textbf{99.32} & 0.49 & 19.09 & 14.89  \\ 

		&&\psmLegend & 2.94 & 2.12 & 1.65 & 1.40 & 0.19 & 0.61 & 11.59 & 16.73 & 84.38 & 97.25 & 98.95 & \textbf{0.61} & 17.40 & 18.69  \\ 

		&&\sgmLegend & 3.13 & 2.03 & 1.73 & 1.40 & 0.23 & 1.03 & 16.08 & 22.98 & 79.80 & 93.94 & 97.22 & 0.26 & 17.02 & 12.76  \\ 

		&&\sparseLegendFull & 3.03 & 2.03 & 1.60 & 1.29 & 0.25 & 0.52 & 10.77 & 23.38 & 84.81 & 94.71 & 97.14 & 0.47 & 16.46 & \textbf{11.57} \\ 

		&&\monodepthLegendFt & 7.02 & 3.17 & 4.48 & 2.54 & 0.43 & 5.09 & 41.45 & 37.06 & 59.26 & 79.24 & 88.39 & 0.41 & \textbf{11.47} & 16.31  \\ 

		\cmidrule{2-17}
		& \multirow{5}{*}{\rotatebox[origin=l]{90}{\parbox[c]{2cm}{\centering \textbf{\small{binned}}}}}
		&\lidarLegend & \textbf{2.42} & \textbf{1.91} & \textbf{1.50} & \textbf{1.31} & \textbf{0.24} & \textbf{1.07} & \textbf{24.30} & \textbf{15.13} & \textbf{81.95} & \textbf{88.92} & \textbf{91.36} & \textbf{0.54} & 14.74 & 4.07  \\ 

		&&\psmLegend & 4.11 & 2.78 & 3.39 & 2.50 & 0.37 & 5.97 & 58.18 & 16.14 & 64.59 & 84.89 & 88.47 & 0.53 & 10.93 & 0.15  \\ 

		&&\sgmLegend & 3.46 & 2.51 & 2.50 & 2.05 & 0.33 & 2.85 & 44.01 & 17.72 & 69.45 & 85.89 & 89.62 & 0.28 & 12.89 & 2.42  \\ 

		&&\sparseLegendFull & 3.80 & 2.61 & 2.89 & 2.24 & 0.38 & 1.75 & 33.09 & 23.88 & 66.31 & 82.34 & 86.45 & 0.44 & 11.77 & 0.95  \\ 
	
		&&\monodepthLegendFt & 9.27 & 3.92 & 7.82 & 3.61 & 0.63 & 33.34 & 154.51 & 33.14 & 41.66 & 65.87 & 78.08 & 0.39 & \textbf{5.85} & \textbf{-4.42}  \\ 
				
		\bottomrule
	\end{tabular}

	}
	\vspace*{-5pt}
	\caption{Quantitative comparison of all benchmarked algorithms based on a variety of 14 metrics averaged over all four scenarios. Although binned results show slightly worse performance, it provides a fairer comparison as depth is not equally distributed.}
	\label{tab:quantitative_results_clear}
	\vspace*{-15pt}
\end{table*}

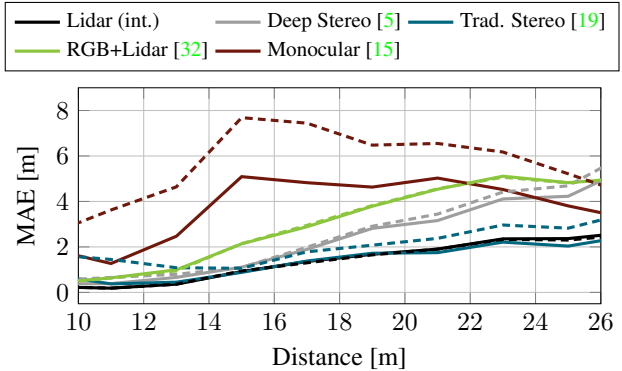
\begin{figure}[t]
	\begin{subfigure}{\columnwidth}
		\centering
		\begin{tikzpicture} 
		\begin{axis}[%
		hide axis,
		xmin=10,
		xmax=50,
		ymin=0,
		ymax=0.4,
		width=5cm,
		legend style={font=\footnotesize, legend cell align=left, legend},
		legend columns=3,
		]
		\addlegendimage{\lidarColor, solid, very thick}
		\addlegendentry{\lidarLegend};
		
		\addlegendimage{\psmColor, solid, very thick}
		\addlegendentry{\psmLegend};
		
		\addlegendimage{\sgmColor, solid, very thick}
		\addlegendentry{\sgmLegend};
		
		\addlegendimage{\sparseColor, solid, very thick}
		\addlegendentry{\sparseLegendFull};
		
		\addlegendimage{\monodepthColor, solid, very thick}
		\addlegendentry{\monodepthLegendFt};
		
		\end{axis}
		\end{tikzpicture}
		\vspace*{5pt}
	\end{subfigure}
	\hfill
\begin{subfigure}{\columnwidth}
\DepthMetric{all}{3}{ylabel=\ac{MAE} {[m]},
					ymin=-0.5, ymax=9,
					scale only axis,
					height=0.13\textheight,
					width=0.85\columnwidth,}{\legend{};}
\end{subfigure}
\vspace*{-20pt}
\caption{\ac{MAE} calculated over depth bins of approximately \unit[2]{m} averaged over all scenarios. Solid lines show daylight performance while dashed lines represent night conditions.}
\label{fig:depth_dependent_metrics_clear}
%\vspace*{-10pt}
\end{figure}

\begin{figure}[t]
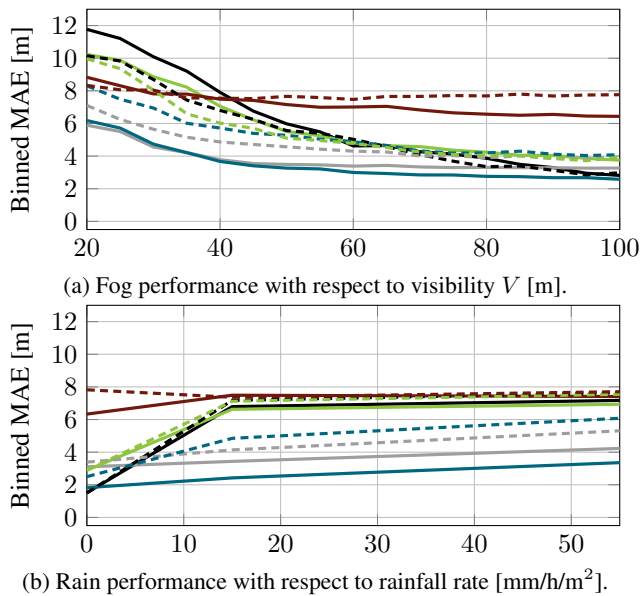

\begin{subfigure}{\columnwidth}
	\MethodComparisonErrorMetric{3}{all}{fog}{_binned}{
	scale only axis,
		xmin=20, xmax=100,
		ymin=-0.5, ymax=13,
		ytick={0,2,4,6,8,10,12},
		height=0.13\textheight,
		width=0.85\columnwidth,
		legend pos=north east,
		ylabel=Binned \ac{MAE} {[m]} ,
		}{\legend{};}
\vspace*{-15pt}
\subcaption{Fog performance with respect to visibility $V$ {[\unit{m}]}.}
\label{fig:fog_rain_performance_binned_fog}
\end{subfigure}
\hfill
\begin{subfigure}{\columnwidth}
	\MethodComparisonErrorMetric{3}{all}{rain}{_binned}{
	scale only axis,
	xmin=0, xmax=55,
	ymin=-0.5, ymax=13,
	ytick={0,2,4,6,8,10,12},
	height=0.13\textheight,
	width=0.85\columnwidth,
	legend pos=north east,
	ylabel=Binned \ac{MAE} {[m]},
	}{\legend{};}
	\vspace*{-15pt}
	\subcaption{Rain performance with respect to rainfall rate {[\unit{mm/h/$\text{m}^2$}]}.}
	\label{fig:fog_rain_performance_binned_rain}
\end{subfigure}
\vspace*{-5pt}
\caption{Binned \ac{MAE} with respect to visibility and rainfall rate. Solid lines show daylight performance while dashed lines represent night conditions. For legend, see Figure~\ref{fig:depth_dependent_metrics_clear}.}
\label{fig:fog_rain_performance_binned}
\vspace*{-15pt}
\end{figure}

First, we describe the evaluation results of all benchmarked algorithms in clear weather conditions. 
Figure~\ref{fig:qualitative_results_clear} shows the 2.5D depth maps, the top-down views and error maps as described in Section~\ref{sec:proposed_benchmark} for the residential area scenario at daylight and night in clear conditions. 
Further qualitative results are shown in the supplemental document.
While the stereo and depth completion approaches produce wrong depth estimates at the edges of objects, monocular depth estimation shows by far worst performance and does not generalize to our data.
The quantitative metrics are averaged over all scenarios and can be found in Table~\ref{tab:quantitative_results_clear}.
This evaluation shows that the monocular and stereo methods slightly decrease in performance at night, while lidar only and lidar depth completion perform stable.
As described in Section~\ref{sec:proposed_benchmark}, these metrics are often calculated over the whole image without considering the depth distribution, and, therefore, we also show a binned variant of the metrics.
The performance slightly decreases as measurements from different distances are now equally weighted, which represents long-range driving scenarios.
In Figure~\ref{fig:depth_dependent_metrics_clear}, \ac{MAE} is plotted with respect to depth. 
Comparing traditional stereo \cite{Hirschmuller2008} with deep stereo \cite{Chang2018}, Table~\ref{tab:quantitative_results_clear} shows that deep stereo methods have difficulties to estimate the correct scale:
Figure~\ref{fig:qualitative_results_clear} shows that the distance to the back wall is estimated approximately \unit[4]{m} closer than what it is.
Nevertheless, for a scale-invariant metric such as \ac{SIlog}, deep stereo \cite{Chang2018} performs better than traditional stereo \cite{Hirschmuller2008}, especially at night, matching the qualitative results in Figure~\ref{fig:qualitative_results_clear}.

%-------------------------------------------------------------------------
\vspace{-9pt}
\paragraph{Evaluation in Fog}

\begin{figure*}[ht!]
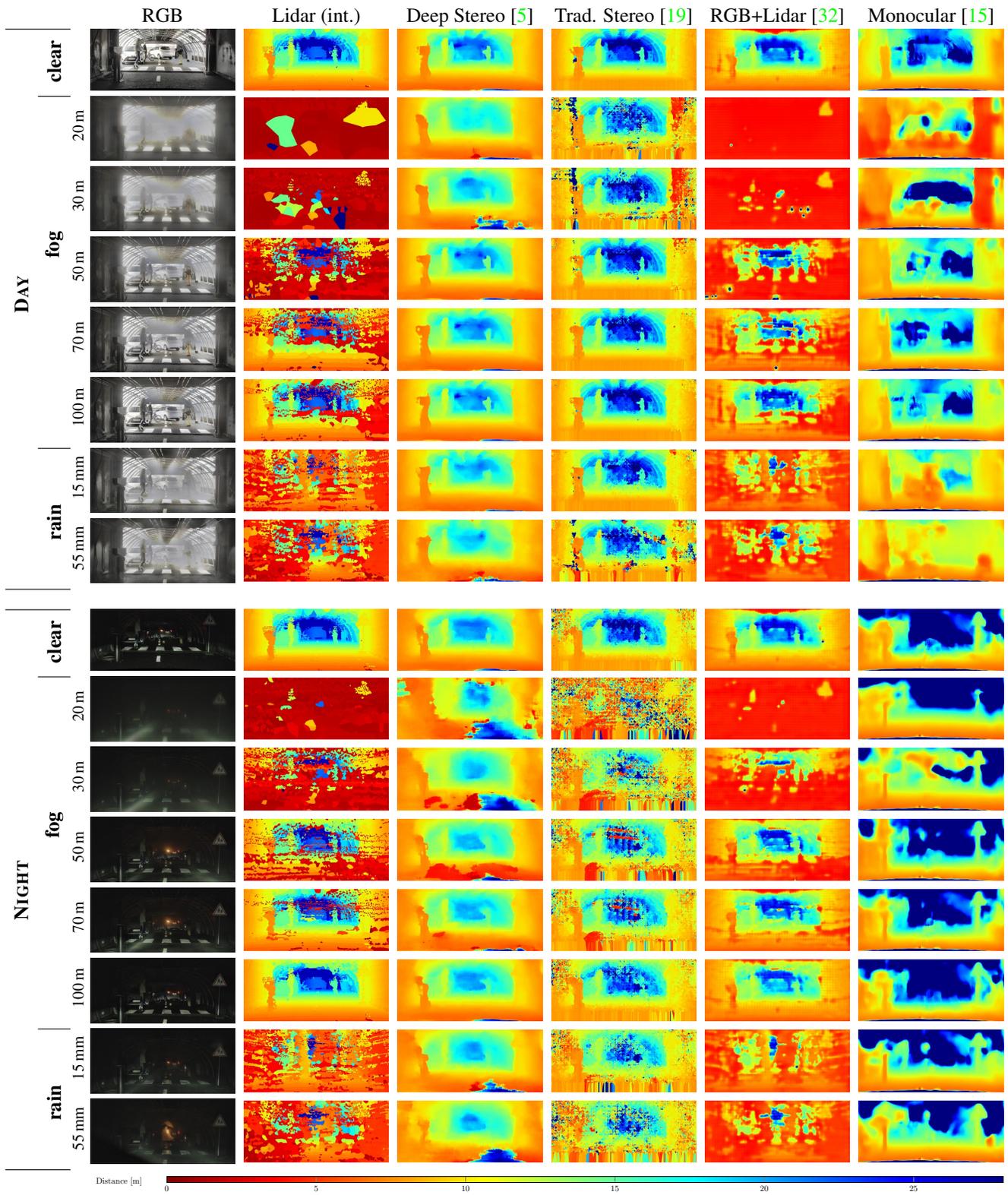

\centering
\qualitativeImagesAll{scene2}
\caption{Qualitative results for all benchmarked depth estimation methods in clear conditions, foggy conditions of different visibility levels and rainy conditions. In addition to the resulting 2.5D depth maps, we include a RGB image as reference.}
\label{fig:qualitative_results_fog_rain}
\end{figure*}

Next, we show how depth estimation algorithms degrade with increasing levels of fog. 
We have extracted images at 17 different visibility levels and show the depth estimation performance with increasing visibility in Figure~\ref{fig:qualitative_results_fog_rain}.
Figure~\ref{fig:fog_rain_performance_binned_fog} shows the \ac{MAE} for varying fog densities.
While contrast in the camera stream drops with decreasing visibility, many cluttered points appear in the lidar point cloud due to severe back-scatter. 
As stereo is based on finding correspondences, the performance of both stereo approaches decreases only slightly with lower visibility. 
In contrast, all methods based on lidar data collapse in the presence of fog because cluttered points cannot be distinguished from measurements originating from ballistic photons, see also \cite{bijelic2018lidar}. 
The RGB input for lidar depth completion only slightly improves the depth recovery performance.
This performance gap can be quantified by comparing clear conditions in daylight with visibility $V=\unit[50]{m}$: between these two settings, deep stereo drops only by \unit[45]{cm} in \ac{MAE}, traditional stereo drops by \unit[84]{cm}, while the \ac{MAE} for lidar-based depth completion reduces by a margin of \unit[2.47]{m} and interpolated lidar even by \unit[5.17]{m}.
Comparing night and daylight conditions, Figure~\ref{fig:fog_rain_performance_binned_fog} indicates that camera-based depth estimation performs worse during night due to dark low-signal images, but lidar-based depth estimation shows slightly better performance during night.
Additional metric visualizations in foggy conditions can be found in the supplemental document.

%-------------------------------------------------------------------------
\vspace{-9pt}
\paragraph{Evaluation in Rain}

We have recorded the scenes at two different rain intensity levels, that is \unit[15]{mm/h/$\text{m}^2$} and \unit[55]{mm/h/$\text{m}^2$}. Figure~\ref{fig:qualitative_results_fog_rain} shows the qualitative impressions while Figure~\ref{fig:fog_rain_performance_binned_rain} visualizes the binned \ac{MAE} corresponding to different intensities of rain. 
We note that light rain disturbs lidar already significantly due to path bending and back-scatter at the droplets while both stereo methods are relatively stable.

%------------------------------------------------------------------------
\section{Conclusion}
In this work, we provide a high-resolution ground truth depth dataset of angular resolution \arcsec{25} for representative automotive scenarios in various adverse weather conditions. Using this evaluation data, we present a comprehensive comparison of state-of-the-art algorithms for stereo depth, depth from mono, lidar and sparse depth completion, in reproducible, finely adjusted adverse weather situations. The sensor data for these comparisons has been acquired with a representative automotive test vehicle.

We demonstrate that stereo approaches robustly generalize to the provided depth dataset, whereas monocular depth estimation is very sensitive to a change of scene type and capture conditions. We also evaluate depth estimation in the presence of fog and rain. We find that stereo-based approaches perform significantly more stable in adverse weather than monocular or recent lidar sensing and depth completion methods based on lidar measurements, which fail due to back-scatter in fog and rain.

All sensor data for the proposed benchmark (1,600 samples) will be made publicly available, and researchers will be able to evaluate their methods based on the high-resolution ground truth data. By acquiring data in a weather chamber, the proposed benchmark allows for reproducible evaluation in adverse weather -- in contrast to existing driving datasets which lack data in adverse weather as current sensing systems fail in these scenarios. As such, this benchmark allows to jointly assess the robustness and resolution of existing depth estimation methods, which is essential for safety in autonomous driving.

%------------------------------------------------------------------------

\vspace*{2mm}
\noindent\small{This work has received funding from the European Union under the H2020 ECSEL Programme as part of the DENSE project, contract number 692449. We thank Robert Böhler, Florian Kraus, Stefanie Walz and Yao Wang for help recording and processing the dataset.}

{\small
\bibliographystyle{ieee}
\bibliography{main}
}

\end{document}